\documentclass{modelica}
\hypersetup{%
  pdftitle  = {A Singularity Avoiding Track Model for Race Car Optimization},
  pdfauthor = {TrackJoint authors},
  pdfsubject = {North American Modelica and FMI Conference 2026},
  pdfkeywords = {Modelica, FMI, race car, track modelling, Monge patch, minimum curvature, Frenet, singularity, optimal control},
  colorlinks,
  linkcolor=black,
  urlcolor=black,
  citecolor=black,
  pdfpagelayout = SinglePage,
  pdfcreator = pdflatex,
  pdfproducer = pdflatex}

\usepackage[utf8]{inputenc}
\usepackage{amsmath}
\usepackage[varbb]{newtxmath} 
\usepackage{bm}
\usepackage{tikz}
\usetikzlibrary{arrows.meta,calc,backgrounds}

\begin{document}
\thispagestyle{empty}

\title{A Singularity Avoiding Track Model for Race Car Optimization}
\author[1]{Adwait Verulkar}
\author[1]{Behzad Danaei}
\author[1]{Srikanth Sivaramakrishnan}
\author[1]{Ryan Trinter}
\affil[1]{General Motors, USA, {\small\texttt{\{adwait.verulkar,behzad.danaei, srikanth.sivaramakrishnan,ryan.trinter\}@gm.com}}}

\maketitle\thispagestyle{empty}

\abstract{%
A track fitting pipeline is presented that converts a measured racetrack point
cloud into a closed, $C^4$ differentiable, three-dimensional
surface model suitable for high-fidelity Modelica race-car optimization.
The surface is a Monge patch,
parametrised by a curvilinear arc-length coordinate measured along the
minimum-curvature line (MCL) and a signed lateral offset, with a height
polynomial of user-selectable degree capturing the
cross-sectional elevation profile.
The choice of MCL as the reference line is deliberate since anchoring the
same surface on the geometric centre-line can cause adjacent normals to
cross inside the drivable region on tight corners. The resulting
non-injective parametrisation assigns multiple curvilinear addresses to the
same three-dimensional point and returns invalid road geometry.
The MCL is obtained from a quadratic program that minimises integrated curvature subject to track-boundary, periodicity,
and jerk-regularisation constraints. By reducing total curvature while
remaining within the track bounds, the MCL moves the region of parametrisation
breakdown outside the drivable strip on all circuits studied, keeps the normal
map injective across the full track width, and yields a more linear
time-to-arc-length transformation for minimum lap-time problems.
The fitted surface and the associated track widths are exported as periodic
piecewise-quintic $C^4$ B-splines, solved as a nonlinear program. The exported
splines are consumed directly by \texttt{TrackJoint}, a Modelica MultiBody
component that enforces the two degree-of-freedom kinematic road constraint via
d'Alembert's principle. We also describe the parts of that Modelica
implementation that set the export format: how the splines are stored on a
shared knot vector as tunable parameters, how a segment is found in constant
time and evaluated in Horner form, why index reduction forces $C^4$
continuity, how the chassis is stacked relative to the track frame, and how the
FMI~3.0 export is used by the CasADi/IPOPT toolchain. A companion paper
\cite{Sivaramakrishnan:2026joint} derives the joint in full.
}


\section{Introduction}

The track model is a crucial component of minimum lap-time simulations of high-performance vehicles. The contact-patch normal load, the gravitational
projections in the body-fixed frame and the kinematic constraints that
keep all four wheels on the road are all expressed in terms of the road
surface and its first two derivatives \cite{LimebeerVanDeMerwe:2025,%
Limebeer:2023nascar}. Inaccuracies in the track model therefore propagate directly into the
predicted lap time, tyre forces, suspension loads and aerodynamic
behaviour, degrading the fidelity of the entire simulation.

The community has converged on a curvilinear $(s,n)$ description for
roads since the work of \textcite{Cossalter:1999}; the recent survey by
\textcite{LimebeerVanDeMerwe:2025} summarises the line of development.
Three-dimensional refinements followed: the Frenet--Serret ribbon of
\textcite{PerantoniLimebeer:2015} parametrises the road by an
arc-length spine plus an orthonormal moving frame and works well on
typical racing layouts but assumes lateral flatness;
\textcite{Lovato:2021} and \textcite{Limebeer:2023nascar} remove that
assumption by superimposing a Monge-patch elevation correction on top
of a Frenet ribbon; and most recently
\textcite{LimebeerVanDeMerwe:2025} dispense with the ribbon altogether
and obtain the surface as a free Monge-patch parametrisation
$\bm{r}_b(s,w)=s\,\bm e^g_s+w\,\bm e^g_w+z(s,w)\,\bm e^g_z$
identified through an optimal control problem on the lateral
polynomial coefficients. The Monge patch removes the need for
Euler angles and yields the first and second fundamental forms in closed
form from the partials of $z(s,w)$, but it leaves open the choice of the
reference line that anchors the $(s,w)$ frame.

\subsection*{The Frenet singularity}

In every published instance of the curvilinear description, including
\cite{LimebeerVanDeMerwe:2025}, the reference line is taken as the
geometric \emph{centre-line} of the track, i.e.\ the average of the
left and right boundary polylines. While this choice is intuitive and
requires no optimisation, it has a structural defect that is fatal for
a production track model, also noted by \textcite{Fork:2024}. Recall that the curvilinear
description maps the planar parameters $(s,w)$ to a 3D point through
\begin{equation}
\bm r_b(s,w) = \bm r(s) + w\,\bm n_{\!g}(s) + z(s,w)\,\bm e_z,
\label{eq:framemap}
\end{equation}
where $\bm r(s)$ is the reference polyline traced at arc length $s$,
$\bm t(s)=\bm r'(s)$ is its unit tangent and
$\bm n_{\!g}(s)=R_{\pi/2}\bm t(s)$ its unit left normal.
\begin{figure}[t]
\centering
\begin{tikzpicture}[
    x={(-0.92cm,0.40cm)},  
    y={(-1.15cm,-0.22cm)}, 
    z={(0cm,1.05cm)},      
    >={Stealth[length=2.2mm]},
    line cap=round,line join=round,
    scale=0.82,
    every node/.style={font=\small}]

  \def\A{0.7}\def\B{0.9}\def\cc{1.5547}\def\hh{1.2}\def\mfac{0.45}
  \def\Arg{(\B*((\t)-\cc))}
  \def\Cy{\A*sin(\Arg r)}                 
  \def\Cyp{\A*\B*cos(\Arg r)}             
  \def\Nrm{sqrt(1+(\Cyp)*(\Cyp))}         
  \def\Mcl{\mfac*\A*sin(\Arg r)}          
  \def\sO{3.3}                            
  \def\ymclO{0.315}                       
  \def\zpar(#1){1.7 + 0.13*(#1) + 0.28*(#1)*(#1)}  
  \def\Tu{(max(0,min(1,((\t)-\sO)/1.5)))}        
  \def\Wfac{(1-0.18*(\Tu)*(\Tu)*(3-2*(\Tu)))}    

  \fill[blue!7,opacity=0.8]
        plot[domain=0.1:6.2,samples=70,variable=\t]
            ({\t-\hh*(\Wfac)*(\Cyp)/(\Nrm)},%
             {(\Cy)+\hh*(\Wfac)/(\Nrm)},0)
     -- plot[domain=6.2:0.1,samples=70,variable=\t]
            ({\t+\hh*(\Wfac)*(\Cyp)/(\Nrm)},%
             {(\Cy)-\hh*(\Wfac)/(\Nrm)},0) -- cycle;

  \draw[gray,thick]
        plot[domain=0.1:6.2,samples=70,variable=\t]
            ({\t-\hh*(\Wfac)*(\Cyp)/(\Nrm)},%
             {(\Cy)+\hh*(\Wfac)/(\Nrm)},0);
  \draw[gray,thick]
        plot[domain=0.1:6.2,samples=70,variable=\t]
            ({\t+\hh*(\Wfac)*(\Cyp)/(\Nrm)},%
             {(\Cy)-\hh*(\Wfac)/(\Nrm)},0);

  \draw[gray!75,dashed,thick]
        plot[domain=0.1:6.2,samples=70,variable=\t] (\t,{\Cy},0);

  \draw[thick,black]
        plot[domain=0.1:6.2,samples=80,variable=\t] (\t,{\Mcl},0);

  \fill[red!7,opacity=0.55]
        (\sO,-0.5,0) -- (\sO,1.9,0) -- (\sO,1.9,3.2) -- (\sO,-0.5,3.2) -- cycle;
  \draw[red!45]
        (\sO,-0.5,0) -- (\sO,1.9,0) -- (\sO,1.9,3.2) -- (\sO,-0.5,3.2) -- cycle;

  \fill (\sO,\ymclO,0) circle (1.3pt);
  \draw[->,thick,black!70] (\sO,\ymclO,0) -- (\sO+1.0,\ymclO,0) node[anchor=south,inner sep=2pt,xshift=19pt,yshift=-7pt] {$\bm t(s)$};

  \draw[->] (\sO,-0.5,0) -- (\sO,1.5,0)
        node[pos=0.5,anchor=north,inner sep=2pt] {$w\,\bm n_{\!g}(s)$};

  \draw[very thick,red,smooth,samples=60,domain=-0.815:1.585,variable=\w]
        plot (\sO,{\ymclO+\w},{\zpar(\w)});

  \draw[->,red] (\sO,1.5,0) -- (\sO,1.5,{\zpar(1.185)})
        node[pos=0.675,anchor=west,inner sep=3pt] {$z(s,w)\,\bm e_z$};
  \fill (\sO,1.5,0) circle (1.3pt);
  \fill[red] (\sO,1.5,{\zpar(1.185)}) circle (1.4pt);
  \node[red,anchor=south west,inner sep=2pt]
        at (\sO,1.5,{\zpar(1.185)}) {$\bm r_b(s,w)$};

  \draw[->,thick,black]
        plot[domain=0.35:1.45,samples=24,variable=\t] (\t,{\Mcl},0);
  \node[anchor=north,inner sep=2pt] at (1.2,{\mfac*\A*sin((\B*(1.2-\cc)) r)},0) {$s$};

\end{tikzpicture}
\caption{Three-dimensional schematic of the Monge-patch track model
$\bm r_b(s,w)=\bm r(s)+w\,\bm n_{\!g}(s)+z(s,w)\,\bm e_z$. The drivable
corridor is delimited by the left/right track boundaries (grey); the
geometric centre-line (dashed) runs midway between them, whereas the
minimum-curvature reference line $\bm r(s)$ (black) is free to move
laterally within the bounds. The
reference-line frame $(\bm n_{\!g}(s),\bm e_z)$ is drawn at a station
along the line. At the marked station~$s$ the vertical cross-section
plane is spanned by the lateral normal $\bm n_{\!g}(s)$ (the $w$ axis)
and $\bm e_z$; the road altitude along that section is the lateral
polynomial $z(s,w)=\sum_{k=0}^{N}z_k(s)\,w^{k}$ (red) shows that the elevation of every surface point $(s,w)$ above the datum plane
equals the value of that polynomial.}
\label{fig:monge-schematic}
\end{figure}
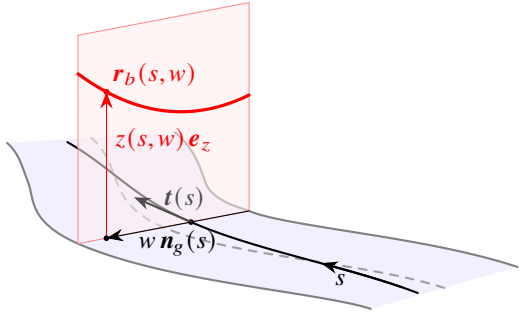

The Frenet--Serret relations on the planar reference give
$\bm t'(s)=\kappa_c(s)\,\bm n_{\!g}(s)$ and
$\bm n_{\!g}'(s)=-\kappa_c(s)\,\bm t(s)$, with $\kappa_c(s)$ the signed
curvature of the reference line. These follow directly from the
tangent-angle parametrisation: writing
$\bm t(s)=(\cos\theta(s),\sin\theta(s))$ with $\theta(s)$ the heading
of the reference tangent and $\kappa_c(s)=\theta'(s)$, the left normal
is $\bm n_{\!g}(s)=R_{\pi/2}\bm t(s)=(-\sin\theta,\cos\theta)$, so that
\begin{align}
\bm t'(s)
 &= \theta'(s)\,(-\sin\theta,\cos\theta)
  = \kappa_c(s)\,\bm n_{\!g}(s), \\
\bm n_{\!g}'(s)
 &= \theta'(s)\,(-\cos\theta,-\sin\theta)
  = -\kappa_c(s)\,\bm t(s).
\end{align}
Figure~\ref{fig:monge-schematic}
illustrates this Monge-patch construction, contrasting the geometric
centre-line with the minimum-curvature reference line introduced below.
Differentiating
\eqref{eq:framemap} yields the in-surface tangent vectors
\begin{align}
\partial_s\bm r_b
 &= \bigl(1-\kappa_c(s)\,w\bigr)\bm t(s)+\partial_s z\,\bm e_z, \\
\partial_w\bm r_b
 &= \bm n_{\!g}(s)+\partial_w z\,\bm e_z,
\end{align}
so that the first fundamental form of the Monge patch,
\begin{equation}
I_P = [\partial_s\bm r_b,\,\partial_w\bm r_b]^{\!\top}
       [\partial_s\bm r_b,\,\partial_w\bm r_b],
\label{eq:firstform}
\end{equation}
has Jacobian determinant
\begin{align}
\sqrt{\det I_P}
 &= \bigl\Vert\partial_s\bm r_b\times\partial_w\bm r_b\bigr\Vert
                                                     \notag\\
 &= \sqrt{\bigl(1-\kappa_c(s)\,w\bigr)^2\bigl(1+(\partial_w z)^2\bigr)
          +(\partial_s z)^2}                          \notag\\
 &\approx \bigl|1-\kappa_c(s)\,w\bigr|\,
    \sqrt{1+\Vert\nabla z\Vert^2}.
\label{eq:jacobian}
\end{align}
where the longitudinal slope $\partial_s z$ is negligible compared
with the lateral slope on racetrack surfaces, so that the vanishing of
the Jacobian is controlled entirely by the factor $1-\kappa_c(s)\,w$.
The map \eqref{eq:framemap} is a local diffeomorphism only where
$\sqrt{\det I_P}>0$, i.e.\ where $1-\kappa_c(s)\,w\neq 0$. The
critical value
\begin{equation}
w_{\mathrm{fold}}(s) = \frac{1}{\kappa_c(s)}
\label{eq:fold}
\end{equation}
is exactly the centre of curvature of $\bm r(s)$ at station $s$.

\begin{figure}[t]
\centering
\includegraphics[width=0.5\textwidth]{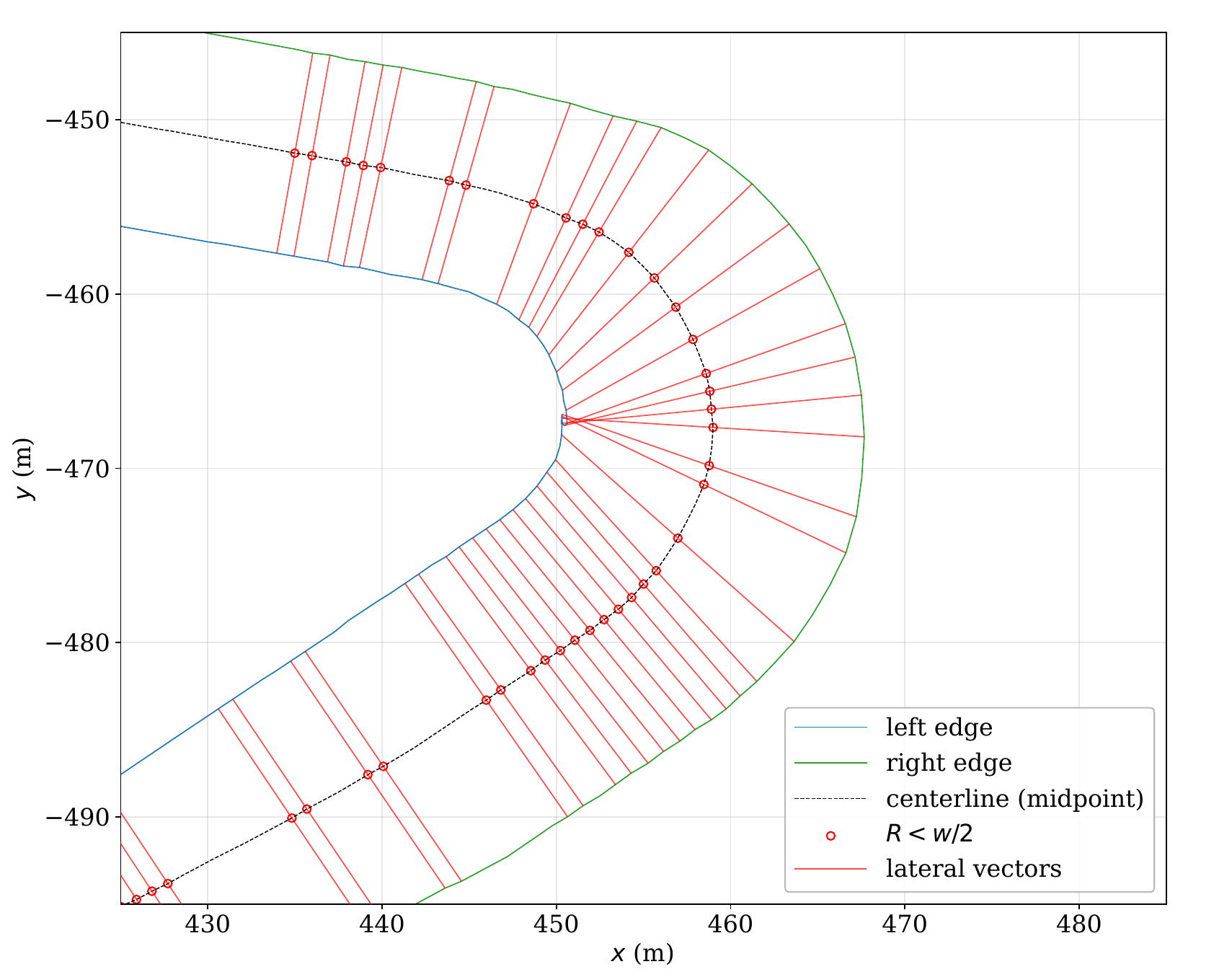}
\caption{Geometric origin of the centre-line defect. Where the local
radius of curvature $1/\kappa_c(s)$ of the reference line falls below
the half-width $b(s)/2$ of the road, neighbouring lateral vectors
$\bm n_{\!g}(s)$ intersect inside the drivable region and the
Frenet map $(s,w)\mapsto\bm r_b(s,w)$ ceases to be injective.}
\label{fig:centerline-curvature}
\end{figure}

On some racetracks the centre-line, taken as a planar polyline
between two surveyed boundaries, has stretches where its local radius
of curvature $1/|\kappa_c(s)|$ is smaller than the local half-width
$b(s)/2$ of the road, so that $|w_{\mathrm{fold}}(s)|<b(s)/2$ and the
fold sits \emph{inside} the drivable region. At every such station
the lateral coordinate frame defined by $\bm n_{\!g}(s)$ is no longer
injective on the full width: two stations $s_1\neq s_2$ produce
laterals that intersect inside the road, as shown in
Figure~\ref{fig:centerline-curvature}, and the Jacobian
\eqref{eq:jacobian} vanishes at $w=w_{\mathrm{fold}}$ and changes
sign beyond it. A vehicle dynamics model that queries the joint at
such a point receives multi-valued geometry and a vanishing
Jacobian, and the resulting tyre-load equations are ill-posed.
We have observed this failure mode
repeatedly on racing circuits that contain a hairpin or a tight
slow-speed sequence, such as the Singapore circuit of the Formula~1
series that is used as the running example throughout this paper.

The same defect is particularly damaging for minimum lap-time
formulations because they are usually solved in the spatial domain by doing an independent-variable transformation. Following the standard
change of independent variable from time $t$ to arc length $s$ along
the reference line, the dynamics of both the vehicle and the track
model are scaled by the factor $1/\dot s$, with
\begin{equation}
\dot s = \frac{v_x\cos\chi - v_y\sin\chi}{1-\kappa_c(s)\,w},
\label{eq:sdot}
\end{equation}
where $v_x,v_y$ are the longitudinal and lateral velocities in the
vehicle-fixed body frame and $\chi$ the heading relative to the
reference tangent. The denominator
$1-\kappa_c(s)\,w$ is the same Jacobian factor as in
\eqref{eq:jacobian}: as the trajectory approaches the fold
$w=w_{\mathrm{fold}}(s)$, $\dot s\to\infty$ and $1/\dot s\to 0$,
collapsing the time transformation and producing a stiff,
ill-conditioned NLP whose Hessian becomes singular at exactly the
stations the optimiser is most interested in. The minimum-curvature
line reduces the curvature peaks responsible for this fold and, on the
processed tracks, keeps $|1-\kappa(s)\,w|$ bounded away from zero on
the full track strip. This yields a well-conditioned spatial-domain
transcription. Moreover, because the minimum-curvature line typically lies near a track boundary in tight corners, the NLP bounds constrain the vehicle path to have no excursion into the singular region.

A secondary effect compounds the singularity in the same corners:
when $w$ is measured from the centre-line, cambered slow corners
place the largest gradients of $z(s,w)$ at large $|w|$ on the side
 opposite the apex, and the degree-$N$ lateral polynomial expansion of $z(s,w)$
must absorb that asymmetry with larger high-order coefficients,
producing spurious wiggles in the partials $\partial_w z$ and
$\partial_{ww} z$ that ultimately drive the tyre normal load.

\subsection*{Proposed solution}

The primary contribution of this paper is a pipeline that fits a
singularity-avoiding parameterisation of any track surface
discretised by a point cloud, the only requirement being that the
point cloud has unique and sorted indices along the longitudinal and
lateral directions. The key component is the \emph{minimum-curvature}
reference line, computed via a quadratic program based on
\textcite{Heilmeier:2020}. Once the reference line is obtained, the
pipeline fits the Monge patch in lateral polynomials of degree~$N$
and exports an arc-length-parameterised, $C^4$-continuous, periodic
piecewise-quintic representation consumed by Modelica-based race-car
optimisation problems.

Racing-line generation has its own large literature that the proposed
pipeline draws on. Direct minimum-time optimal control
\cite{PerantoniLimebeer:2015} provides the true optimum but is
expensive and requires a vehicle model. Sequential or geometric
approaches such as the minimum-curvature quadratic program of
\textcite{Braghin:2008} and the iterated, constraint-aware
formulation of \textcite{Heilmeier:2020} are within a fraction of a
percent of the time-optimal solution while being orders of magnitude
faster and vehicle-agnostic; the two-step decomposition of
\textcite{Kapania:2016} alternates path and velocity profile. The novel contribution of this paper is to repurpose the
minimum-curvature line of \textcite{Heilmeier:2020} and
\textcite{Braghin:2008} not as a driving target but as the
curvilinear reference that anchors the Monge-patch surface
parametrisation of \textcite{LimebeerVanDeMerwe:2025}, thereby
certifying injectivity of the Frenet map on the full track strip.

The rest of the paper is organised as follows.
Section~\ref{sec:methodology} develops the fitting pipeline, from the
minimum-curvature quadratic program to the $C^4$ piecewise-quintic
export. Section~\ref{sec:trackjoint} shows how the Modelica
\texttt{TrackJoint} component uses those splines, and why the
differential-algebraic equation it forms needs each property of the
export format. Section~\ref{sec:results} reports the effect on the
conditioning of a full-circuit minimum lap-time problem.

\section{Methodology}\label{sec:methodology}

The end-to-end pipeline is organised in seven stages:

\begin{enumerate}
\item Parse the structured point cloud
      $\{(j,k,x_{jk},y_{jk},z_{jk},\bm n_{jk})\}$ exported by the
      surveyor, where $j$ indexes the longitudinal station, $k$ the
      lateral sample at that station, $(x_{jk},y_{jk},z_{jk})$ is the
      measured 3D point and $\bm n_{jk}$ the associated unit surface
      normal;
\item Extract the median-lateral reference polyline and the left/right
      boundaries from the point-cloud lateral indexing;
\item Solve the minimum-curvature quadratic program on the boundary
      polylines, iterating on the local linearisation, with periodic
      $\alpha$-smoothing to suppress active-constraint chatter;
\item Build the Frenet frame on the resulting line and recover the
      lateral widths $b_\mathrm{left}(s)$, $b_\mathrm{right}(s)$ by
      ray-polyline intersection with the boundaries;
\item Re-sample the point cloud onto a longitudinal--lateral grid
      that is orthonormal with respect to the minimum-curvature line,
      i.e.\ aligned with the new $(s,w)$ frame;
\item Fit, at each station, the Monge-patch elevation
      $z(s,w)=\sum_{k=0}^{N}z_k(s)\,w^k$ in the lateral coordinate;
\item Fit each longitudinal channel
      ($r_x$, $r_y$, $z_0,\ldots,z_N$, $b_\mathrm{left}$,
      $b_\mathrm{right}$) by a periodic piecewise quintic with
      $C^4$ continuity at the knots that is consumed by the Modelica-based vehicle and track model.
\end{enumerate}

\subsection{Surface model}\label{sec:surface}

The track is a smooth, regular, orientable two-dimensional manifold
$\mathcal M\subset\mathbb R^3$. We parametrise it by the Monge patch
\eqref{eq:framemap} of \cite{LimebeerVanDeMerwe:2025}, with $(s,w)$ in
a rectangular domain $\mathcal D\subset\mathbb R^2$, $\bm r(s)$ the
reference line at arc length $s$, $\bm n_{\!g}(s)$ its unit left normal
and $\bm e_z$ the vertical unit vector. The
elevation is taken in the algebraic-function class of
\cite{LimebeerVanDeMerwe:2025},
\begin{equation}
z(s,w) = \sum_{k=0}^{N} z_k(s)\,w^k,
\label{eq:zexp}
\end{equation}
truncated at a user-selectable degree~$N$. In practice we adopt the low order $N=3$ proposed by
\textcite{LimebeerVanDeMerwe:2025}, which is sufficient to capture
camber and crown on every circuit we have processed while keeping the
lateral derivatives $\partial_w z$ and $\partial_{ww} z$ well behaved.

For any $N\ge 2$ these lateral derivatives exist, and provided each
coefficient $z_k(s)$ is a $C^4$ piecewise polynomial in $s$ they are also
continuously differentiable in~$s$ up to second order and beyond. Together with
$\partial_s\bm r_b$ and $\partial_w\bm r_b$,
which are linearly independent everywhere by construction, this
provides every quantity needed for the first and second fundamental
forms~$I_P$ and $\Pi_P$:
\begin{align}
I_P &= \begin{bmatrix}
(1-\kappa_c w)^2+(\partial_s z)^2 & \partial_s z\,\partial_w z \\
\partial_s z\,\partial_w z & 1+(\partial_w z)^2
\end{bmatrix}, \\[4pt]
\Pi_P &= \begin{bmatrix}
\bm e_m^{P\top}\partial_{ss}\bm r_b &
\bm e_m^{P\top}\partial_{sw}\bm r_b \\
\bm e_m^{P\top}\partial_{sw}\bm r_b &
\bm e_m^{P\top}\partial_{ww}\bm r_b
\end{bmatrix},
\end{align}
with
$\bm e_m^P=(\partial_s\bm r_b\times\partial_w\bm r_b)/
\Vert\partial_s\bm r_b\times\partial_w\bm r_b\Vert$
the unit normal. Together, $I_P$, $\Pi_P$ and $\bm e_m^P$ provide the
gravity projections, the kinematic constraints involving the body-fixed
angular velocities $\omega_x^b,\omega_y^b$, and the closed-form normal
acceleration $\ddot{\bm r}_b^\top\bm e_m^P$ of
\cite{LimebeerVanDeMerwe:2025}. With positive camber defined by positive
lateral slope in the $w$ direction, the local camber angle is
$\phi(s,w)=\arctan(\partial_w z(s,w))
=\arctan\!\bigl(\sum_{k=1}^{N}k\,z_k(s)\,w^{k-1}\bigr)$.

\subsection{Minimum curvature reference line}\label{sec:minc}

Here we replace the centre line by the
minimum curvature line. The minimum curvature objective minimises the
integral of $\kappa^2(s)$ over the closed loop subject to the bounds of the track. We also
verify non-intersection of the recovered laterals after the line and
widths are recomputed in the minimum-curvature frame.
A further geometric benefit is that the line typically hugs the
inside boundary through slow corners, so the admissible $(s,w)$
excursion is effectively one-sided and the relevant $|w|$ shrinks,
improving the conditioning of both the lateral polynomial fit and
the spatial-domain time transformation $1/\dot s$ of
\eqref{eq:sdot}.

\subsubsection{Quadratic program}\label{sec:qp}

Following \textcite{Braghin:2008} and the iterative
constraint-aware refinement of \textcite{Heilmeier:2020}, candidate
points along the line are written as
\begin{equation}
\bm r_i = \bm p_i + \alpha_i\,\bm n_i,
\qquad \alpha_i\in\bigl[-b_{\mathrm r,i}+\tfrac{b_v}{2},\;
b_{\mathrm l,i}-\tfrac{b_v}{2}\bigr],
\label{eq:alpha}
\end{equation}
where $\bm p_i$ is a reference point (initially the centre-line),
$\bm n_i$ is its unit left normal, $b_{\mathrm l,i},b_{\mathrm r,i}$ the
local left/right widths recovered by ray-polyline intersection against
the boundary polygons, and $b_v$ the vehicle width. The objective is a
weighted sum of squared discrete curvatures with curvature samples
written as second-order finite differences. We retain the cross term
in the discrete curvature expansion and omit explicit curvature
constraints because the closed-loop containment check described below
guarantees strict containment.

The objective is augmented with two regularisers. A length-cost
$w_L\sum_i\Vert\bm r_{i+1}-\bm r_i\Vert^2$ discourages local stretching
of the discretisation when the boundaries are noisy. A jerk-cost
$w_J\sum_i\Vert\bm r_{i+2}-3\bm r_{i+1}+3\bm r_i-\bm r_{i-1}\Vert^2$
removes residual two-sample chatter at stations where the bound
constraints are exactly active. To reduce sensitivity to non-uniform
sampling, the second- and third-difference rows are weighted by
$1/\Delta s_i^{3/2}$ and $1/\Delta s_i^{5/2}$, respectively. The
resulting QP has the standard form
$\min\tfrac12\bm\alpha^\top H\bm\alpha+f^\top\bm\alpha$ subject to
the margin-shrunk box constraints
$-\bm b_{\mathrm r}\le\bm\alpha\le\bm b_{\mathrm l}$. Concretely, the
three penalty groups are assembled into a single weighted
least-squares residual $A\bm\alpha-\bm c$. We introduce the periodic
circulant difference operators $D_1,D_2,D_3$ with stencils
$(-1,1)$, $(1,-2,1)$ and $(-1,3,-3,1)$, the diagonal weight matrices
$W_\kappa=\operatorname{diag}(\Delta s_i^{-3/2})$,
$W_L=\sqrt{w_L}\,\operatorname{diag}(\Delta s_i^{-1/2})$ and
$W_J=\sqrt{w_J}\,\operatorname{diag}(\Delta s_i^{-5/2})$, and the
normal-projection diagonals $N_x=\operatorname{diag}(n_{i,x})$,
$N_y=\operatorname{diag}(n_{i,y})$. With $\bm p_x,\bm p_y$ the
reference-point coordinate vectors, the residual is the block-sparse
stack $A^\top=[\,$ $N_x D_2 W_\kappa$, $N_y D_2 W_\kappa$,
$N_x D_1 W_L$, $N_y D_1 W_L$, $N_x D_3 W_J$, $N_y D_3 W_J$ $\,]$ and
$\bm c^\top=-[\,$ $\bm p_x^\top D_2 W_\kappa$, $\bm p_y^\top D_2 W_\kappa$,
$\bm p_x^\top D_1 W_L$, $\bm p_y^\top D_1 W_L$,
$\bm p_x^\top D_3 W_J$, $\bm p_y^\top D_3 W_J$ $\,]$,
where the $x$ and $y$ component rows are stacked per penalty group.
The Hessian and the linear term of the QP then follow as the
Gauss--Newton products $H = A^\top A$ and $f = -A^\top\bm c$,
so that $H$ is symmetric positive semidefinite and banded by the
four-point periodic stencil, and $\bm c$ collects the curvature,
length and jerk of the current frozen reference $\{\bm p_i,\bm n_i\}$.
The QP is solved with OSQP~\cite{Stellato:2020}.

\begin{figure}[t]
\centering
\includegraphics[width=0.45\textwidth]{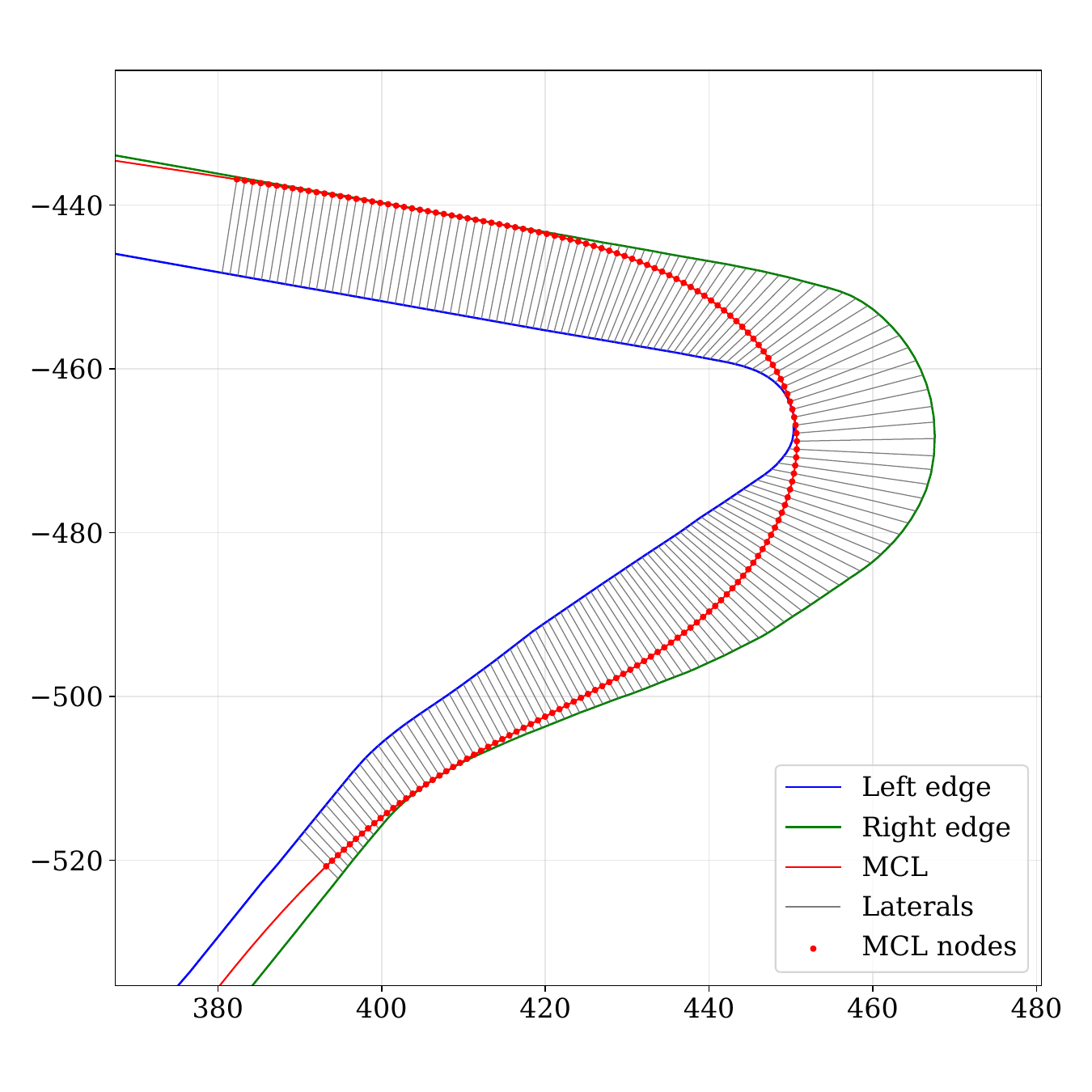}
\caption{Result of the proposed reference-line construction on the
same hairpin region as Fig.~\ref{fig:centerline-curvature}. The
minimum-curvature line (MCL, red) stays strictly inside the track
and the recovered laterals (grey) no longer intersect within the
drivable region, restoring injectivity of the Frenet map.}
\label{fig:laterals-mcl}
\end{figure}

Figure~\ref{fig:laterals-mcl} shows the same hairpin region as
Fig.~\ref{fig:centerline-curvature}, now anchored on the
minimum-curvature line produced by the pipeline of
Section~\ref{sec:minc}. It can be seen that the recovered laterals no longer cross
inside the drivable region.

\subsubsection{Iteration, smoothing and containment}\label{sec:qp-iteration}

A single QP solve does \emph{not} return the true minimum-curvature
line, because the parameterisation $\bm r_i=\bm
p_i+\alpha_i\bm n_i$ in \eqref{eq:alpha} is linear in $\alpha_i$ only
for a \emph{frozen} reference $\{\bm p_i,\bm n_i\}$, whereas the
underlying minimum-curvature problem is nonlinear in the position of
the line itself. Three coupled approximations are introduced when the
reference is fixed:
\begin{enumerate}
\item \textbf{Geometric linearisation of curvature.} The discrete
      curvature operator is built as a second finite difference of
      $\alpha_i$ projected onto the reference normals $\bm n_i$. This
      is a Gauss--Newton-style linearisation of $\kappa\,ds$ about the
      current line; the true $\kappa$ depends nonlinearly on the
      tangent rotation between adjacent samples, which itself changes
      once $\bm\alpha$ moves the line. A single solve therefore only yields
      a curvature-reducing \emph{step}.
\item \textbf{Stale arc-length and normals.} The row weights
      $1/\Delta s_i^{3/2}$ and $1/\Delta s_i^{5/2}$, the normals $\bm
      n_i$ and the second-difference stencil are all computed on the
      \emph{previous} reference. After the line shifts, $\Delta s_i$
      and the local normal direction change, so the discrete operator
      no longer represents $\int\kappa^2\,ds$ on the updated geometry
      until the reference is refreshed and the arc-length resampling
      is repeated.
\item \textbf{Stale lateral bounds.} The half-widths $b_{\mathrm
      l,i},b_{\mathrm r,i}$ are obtained by ray-casting from the
      reference $\bm p_i$ along $\bm n_i$ to the boundary polygons.
      Once the line has moved laterally by $\alpha_i$, the ray
      footpoint and direction at the new station are different and the
      old bounds are merely an approximation; in particular, on tight
      corners the new line can land outside the polygon that the old
      bounds claimed to enclose it in (see the containment step below).
\end{enumerate}
We therefore wrap the QP in a fixed-point iteration:
$\bm p^{(k+1)}_i\leftarrow\bm p^{(k)}_i+\alpha^{(k)}_i\bm n^{(k)}_i$
followed by an arc-length re-sampling and recomputation of normals,
$\Delta s_i$ and ray-cast widths before the next solve. In practice, 3--5 iterations suffice.

This outer loop is structurally an SQP step. It's effectively a convex QP built on a
linearisation of a nonlinear problem, iterated to a fixed point. We avoid a general-purpose SQP/NLP
solver such as IPOPT~\cite{WaechterBiegler:2006} because the dominant
nonlinearities such as ray-polyline intersections for width computation, polygon containment of the MCL and
arc-length resampling, are non-smooth operations in terms of $\bm\alpha$. With these quantities
frozen, each subproblem is a convex QP with banded Hessian $H=A^\top A$,
making OSQP a natural choice. 

Because widths are computed at the previous reference, the new line
can in pathological cases land slightly outside the boundary polygons.
A boolean inside/outside test using
\texttt{contains\_points} flags such points after every solve. If any
point is flagged, the iteration is rejected and the line reverts to the
most recent iteration whose points were all strictly inside. If the bounds become active after the first QP solve, no previous iterate is available to fall back on, so we keep the solution and rely on the smoothing step below to remove bound-induced noise in the line and laterals. After each solve, the discrete $\bm\alpha$ is
filtered with a short periodic centred moving average that suppresses
the two-sample zig-zag arising at stations where the bound constraints
are active.

\subsubsection{Frenet frame and track widths}\label{sec:frenet}

The closed minimum-curvature line is fitted with a periodic quintic
B-spline $(x(s),y(s))$ in arc length. Tangents and unit left normals
follow analytically:
\begin{equation}
\bm t(s)=\frac{(x'(s),y'(s))}{\Vert(x'(s),y'(s))\Vert},\quad
\bm n_{\!g}(s)=R_{\pi/2}\bm t(s).
\end{equation}
Once the reference line is found using the logic of the previous
section, the accurate lateral half-widths
$b_\mathrm{left}(s),b_\mathrm{right}(s)$ are computed by ray-polyline
intersection of the local lateral against both boundaries in both signs,
taking the smallest positive parameter.

\subsection{Re-sampling and Monge-patch fit}\label{sec:fit}

\begin{figure}[t]
\centering
\includegraphics[width=0.4\textwidth]{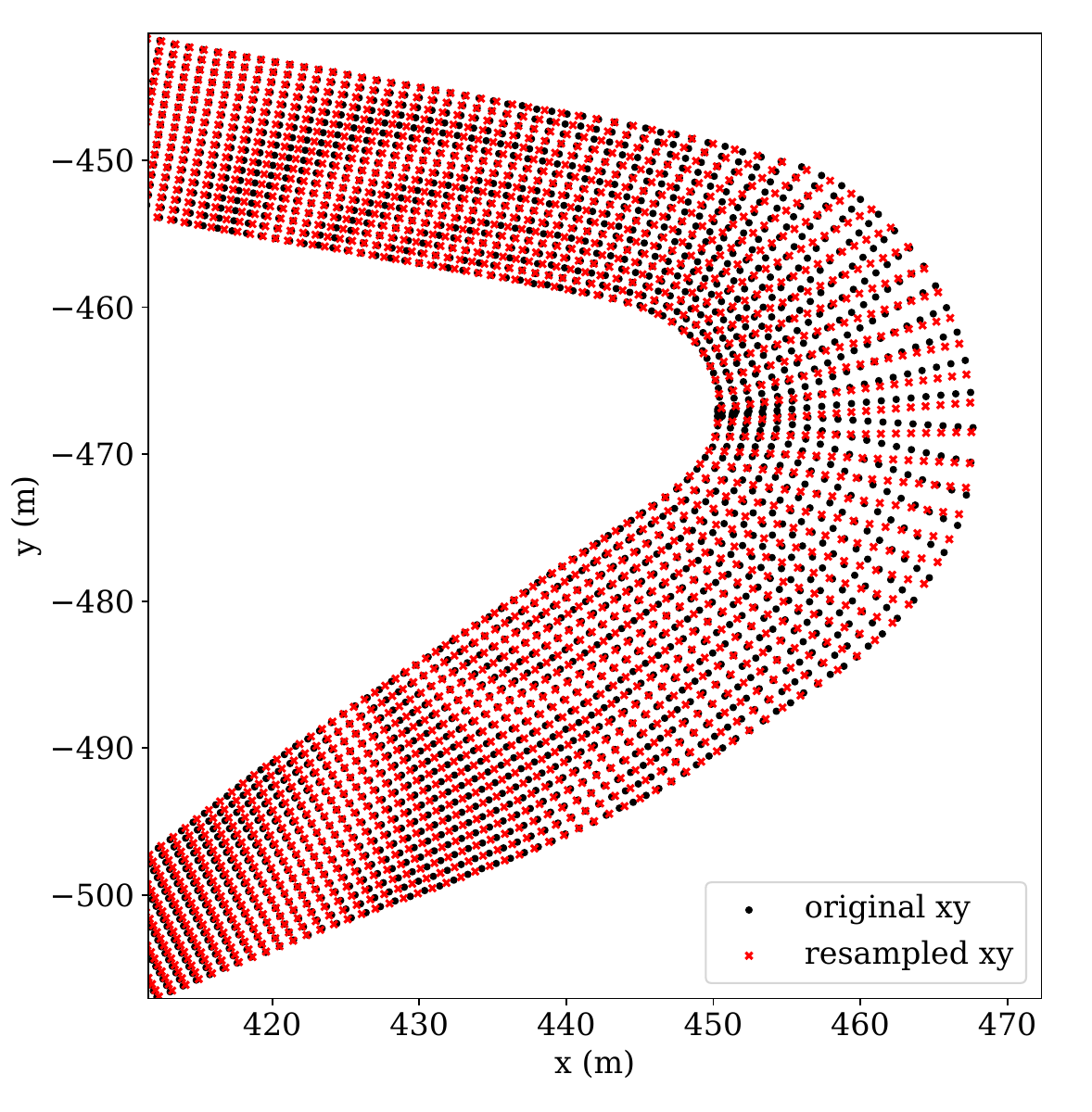}
\caption{Regular $(n_\mathrm{long}\times n_\mathrm{lat})$ resampling
of the surveyed point cloud on the minimum-curvature frame.}
\label{fig:resampled-points}
\end{figure}

\begin{figure*}[t]
\centering
\includegraphics[width=\textwidth]{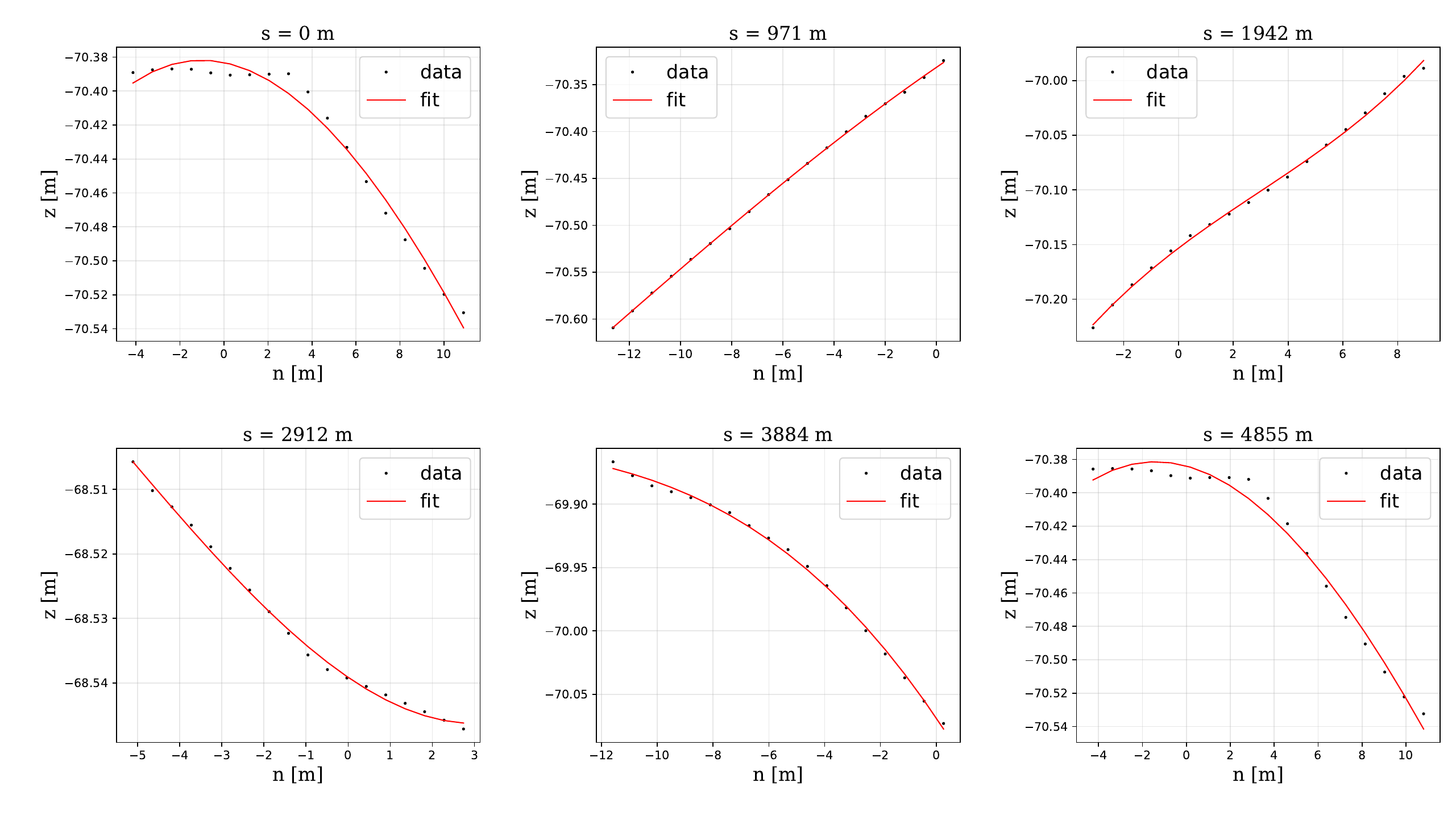}
\caption{Lateral degree-$N$ polynomial fit of the elevation
$z(s_i,w_k)$ at each longitudinal station.}
\label{fig:lateral-fits-per-station}
\end{figure*}

A regular $(n_\mathrm{long}\times n_\mathrm{lat})$ grid is generated
on the minimum-curvature frame, so that the sample stations are
orthonormal with respect to the reference line and lateral stations sweep
$w_k\in[-b_\mathrm{right}(s_i),\,b_\mathrm{left}(s_i)]$, as shown in
Figure~\ref{fig:resampled-points}. The
elevation at every grid point $(s_i,w_k)$ is obtained from an
internal tool that retains a high-density representation of the
surveyed surface and queries it at the grid stations to provide
elevation values at arbitrary $(x,y)$ locations beyond those of the
original samples. The tool uses Delaunay triangulation of the
surveyed points as the primary interpolant, falling back to a
radial-basis-function (thin-plate-spline kernel, 50 nearest
neighbours) interpolant for query points that lie outside the
convex hull of the triangulation. At every longitudinal station the elevation samples
$z(s_i,w_k)$ are fitted by least squares with the degree-$N$ lateral
polynomial of \eqref{eq:zexp} as shown in Figure~\ref{fig:lateral-fits-per-station}.

\subsection{Export: $C^4$ piecewise quintic}\label{sec:export}

The Modelica model consumes a periodic piecewise-quintic
description with shared knots: each of the $N+5$ channels
($r_x,r_y,z_0,\ldots,z_N,b_\mathrm{left},b_\mathrm{right}$) is a
polynomial of degree 5 on each segment with $C^4$ continuity at
interior knots and strict periodicity of the function value and the
first four derivatives at the wrap. Quintics are needed for two
reasons: $z(s,w)$ derivatives must be continuous through the
expressions for $I_P$ and $\Pi_P$ (Section~\ref{sec:surface}), and the
track FMU evaluates these in real time inside a Modelica DAE solver
that is sensitive to discontinuities in slope.

The fit is formulated as a smoothness-regularised periodic
nonlinear program in CasADi~\cite{Andersson:2019} and solved by
IPOPT~\cite{WaechterBiegler:2006} with the HSL MA57 linear
solver~\cite{HSL:MA57}. For a fixed number of segments
$N_\mathrm{seg}$ and a polynomial degree $d=5$, the variables are the
local-basis coefficients
$\{a_{i,j}\}_{i=0,\ldots,N_\mathrm{seg}-1;\,j=0,\ldots,d}$:
\begin{equation}
y_i(s) = \sum_{j=0}^{d} a_{i,j}\,(s-s_i)^j,
\quad s\in[s_i,s_{i+1}].
\end{equation}
The objective combines the per-sample least-squares mismatch and the
integrals, evaluated by 3-point Gauss--Legendre quadrature, of the
squared first to fourth derivatives:
\begin{equation}
\frac1m\sum_{p=1}^{m}\bigl(y_{i(p)}(s_p)-y_p\bigr)^2
+\sum_{q=1}^{4} w_q\!\int_0^{L}\!\!\bigl(y^{(q)}(s)\bigr)^2\,\mathrm ds.
\end{equation}
Hard $C^4$ continuity is imposed at every interior knot and at the
wrap. For the $N+1$ elevation channels we additionally
apply a zero-phase FFT low-pass filter with cutoff wavelength equal to
the segment length to prevent the quintic from being asked to represent
longitudinal waves in vertical height whose wavelength is comparable
to the spacing of its knots, which would otherwise appear as
under-resolved spline oscillations.

\section{Modelica Implementation}\label{sec:trackjoint}

This section describes the receiving end of
the pipeline, the \texttt{TrackJoint} component. The companion paper
\cite{Sivaramakrishnan:2026joint} gives the full derivation of the
joint, its validation against the LiDAR terrain mesh and its use in
time-domain simulation.

\texttt{TrackJoint} is a holonomic two-degree-of-freedom kinematic
joint that realises the Monge patch \eqref{eq:framemap} as a
\texttt{Modelica.Mechanics.MultiBody}~\cite{Otter:2003} component. Its
joint coordinates are the same two curvilinear parameters used
throughout this paper: the arc-length coordinate $s$ (\texttt{sTrack})
measured along the minimum-curvature line, and the signed lateral
offset $w$ (\texttt{wTrack}). From these two coordinates and the
exported channels the joint builds the surface position, the
orientation triad and the angular velocity of a single
\texttt{frame\_b} connector, and closes the road constraint through
d'Alembert's principle.

\subsection{Parameter record and spline storage}\label{sec:record}

Of the $N+5$ exported channels the joint consumes only the $N+3$
geometry channels $\{r_x,r_y,r_z,z_1,\ldots,z_N\}$, where
$r_z\equiv z_0$ of \eqref{eq:zexp}; at the adopted order $N=3$ this is
the six-channel set $\{r_x,r_y,r_z,z_1,z_2,z_3\}$. The two width
channels $b_\mathrm{left},b_\mathrm{right}$ are not part of the
kinematics and are consumed instead as the path constraint
$-b_\mathrm{right}(s)\le w\le b_\mathrm{left}(s)$ on the joint
coordinate \texttt{wTrack} in the optimal-control problem.

All geometry enters the joint through a single parameter record. Each
channel is stored as the matrix of local monomial coefficients
$a_{i,j}$ of Section~\ref{sec:export}, of size
$N_\mathrm{seg}\times(d{+}1)$ with $d=5$, together with one break
vector shared by every channel:

\begin{lstlisting}[caption={Spline storage in the \texttt{TrackJoint}
parameter record: one coefficient matrix per exported channel and a
single shared break vector.},label=lst:record,basicstyle=\ttfamily\footnotesize]
  parameter Real sBreak[nSeg + 1];
  parameter Real rxCoeff[nSeg, d + 1];
  parameter Real ryCoeff[nSeg, d + 1];
  parameter Real rzCoeff[nSeg, d + 1];
  parameter Real z1Coeff[nSeg, d + 1];
  parameter Real z2Coeff[nSeg, d + 1];
  parameter Real z3Coeff[nSeg, d + 1];
\end{lstlisting}

Sharing one break vector across channels is why the export step of
Section~\ref{sec:export} solves all channels on a common knot
sequence. The segment index is then found once per integration step
and reused by every channel. The export also makes the value and the
first four derivatives periodic at the wrap, so the joint needs no
special case at the start/finish line and simply reduces the
arc-length coordinate modulo the track length $L$. The coefficient
matrices are declared \texttt{tunable}, so one compiled FMU can be
pointed at a different circuit at run time without recompiling.

\subsection{Segment lookup and Horner evaluation}

Because the exported breaks are uniform, the segment index at a query
$s$ is obtained in constant time,
\begin{equation}
\mathrm{ix} = \min\!\left(
  \left\lfloor \frac{s \bmod L}{\Delta s} \right\rfloor,\;
  N_\mathrm{seg}-1\right),
\label{eq:ix}
\end{equation}
avoiding the branch-heavy interval search of a general B-spline
evaluator. Every channel derivative is then the local monomial sum
\begin{equation}
y^{(q)}(s) = \sum_{j=q}^{d}
  \frac{j!}{(j-q)!}\,a_{\mathrm{ix},j}\,(s-s_{\mathrm{ix}})^{\,j-q},
\label{eq:splineeval}
\end{equation}
evaluated in Horner form by one Modelica function that takes the
derivative order $q$ as an argument. Both \eqref{eq:ix} and
\eqref{eq:splineeval} are branch-free in the continuous variables, so
the whole joint traces to a smooth expression graph when CasADi
differentiates the FMU at each collocation point.

This is also where the degree and continuity chosen in
Section~\ref{sec:export} come from. The joint needs
$\partial_{sss} r_x$ and $\partial_{sss} r_y$ to form the parametric
rotation rates, so $q$ reaches~3. The Modelica tool then
differentiates the angular-velocity equation once more during index
reduction, which takes the requirement to $q=4$. A piecewise quintic
with $C^4$ continuity at every interior knot and at the wrap is the
lowest degree that keeps everything the DAE solver differentiates
continuous across a knot. A cubic spline leaves
$\partial_{ss}\bm r_b$ jumping at each knot, which makes the
integrator chatter at the segment boundaries.

\subsection{Surface geometry and the d'Alembert constraint}

Listing~\ref{lst:trackjoint} reproduces the equation block that turns
the evaluated channels into surface geometry. The height polynomial
\eqref{eq:zexp} and its partials are expanded in Horner form, the
in-surface tangents $\partial_s\bm r_b$ and $\partial_w\bm r_b$ follow
directly, and the triad $(\bm n_x,\bm n_y,\bm n_z)$ is built
analytically from them, so that no Euler angles or quaternions are
integrated inside the joint.

\begin{lstlisting}[caption={Core surface block of the Modelica
\texttt{TrackJoint} that consumes the exported channels
$\{r_x,r_y,r_z,z_1,z_2,z_3\}$: Horner height polynomial, surface
position and partials, track-frame triad and d'Alembert constraint.
The full component is given in \cite{Sivaramakrishnan:2026joint}.},label=lst:trackjoint,basicstyle=\ttfamily\footnotesize]
  // Height z(s,w) in Horner form and partial derivatives
  z = rz + wTrack*(z1 + wTrack*(z2 + wTrack*z3));
  dzds = drzds + wTrack*(dz1ds + wTrack*(dz2ds + wTrack*dz3ds));
  dzdw = z1 + wTrack*(2*z2 + wTrack*(3*z3));

  // Surface position: lateral offset along ground-plane normal
  rTrack = {rx - wTrack*dryds, ry + wTrack*drxds, z};

  // First partial derivatives of surface position
  drds = {drxds - wTrack*d2ryds2, dryds + wTrack*d2rxds2, dzds};
  drdw = {-dryds, drxds, dzdw};

  // Track frame triad: tangent, surface normal, binormal
  nxTrack = normalize(drds);
  nzTrack = normalize(cross(drds, drdw));
  nyTrack = cross(nzTrack, nxTrack);

  // D'Alembert: virtual work in s and w directions must vanish
  0 = fFrameB*drds + tFrameB*dphids;
  0 = fFrameB*drdw + tFrameB*dphidw;

  // Frame_b assignment: position and rotation matrix
  frame_b.r_0 = rTrack;
  frame_b.R.T = transpose([nxTrack, nyTrack, nzTrack]);
\end{lstlisting}

The two constraint equations in the listing are the joint's link back
to the geometry of Section~\ref{sec:surface}. Instead of introducing
Lagrange multipliers for the directions orthogonal to $\mathcal M$,
the body-frame wrench $(\bm f_b,\bm\tau_b)$ arriving on
\texttt{frame\_b} is resolved to world orientation with
\texttt{resolve1} and the two generalised forces conjugate to $(s,w)$
are set to zero,
\begin{align}
0 &= \bm f^w_b\!\cdot\!\partial_s\bm r_b
    +\bm\tau^w_b\!\cdot\!\partial_s\bm\phi,
\label{eq:dAlembertS}\\
0 &= \bm f^w_b\!\cdot\!\partial_w\bm r_b
    +\bm\tau^w_b\!\cdot\!\partial_w\bm\phi,
\label{eq:dAlembertW}
\end{align}
where $\partial_s\bm\phi,\partial_w\bm\phi$ are the parametric
rotation-rate vectors of the triad. These two relations close the
index-1 DAE in $(\dot s,\dot w)$ and are linear in the wrench
components. Their $2\times 6$ Jacobian has rank two wherever the
parametrisation is regular, and the Frenet fold \eqref{eq:fold} is the
configuration in which it loses rank. So the minimum-curvature
reference line of Section~\ref{sec:minc} does more than improve the
conditioning of the NLP. It is what keeps the Modelica constraint
solvable at all.

\subsection{State selection and MultiBody assembly}

The joint declares $(s,\dot s,w,\dot w)$ as
\texttt{StateSelect.always} variables so that the tool keeps them as
continuous states whatever the surrounding vehicle topology. Without
this, state aliasing against a chassis that is otherwise free in
$\bm e_z$ makes the dynamic state selection fail. A
\texttt{Connections.root(frame\_b.R)} statement makes the joint the
root of the orientation graph of the subtree it heads, following the
pattern of the standard MultiBody joints.

In a vehicle assembly the joint never carries the chassis alone.
Stacked on \texttt{frame\_b} are a standard
\texttt{Modelica.\allowbreak Mechanics.\allowbreak MultiBody.\allowbreak Joints.\allowbreak Prismatic}
heave joint along the surface binormal $\bm n_z$ and three revolute
joints realising roll, pitch and yaw \emph{relative to the track
triad}. The chassis thus keeps all six degrees of freedom, but each
one is expressed relative to the road rather than to the world frame.

\subsection{FMI 3.0 export and use in FMIOPT}\label{sec:fmiexport}

The joint and the host MultiBody vehicle model are compiled in Dymola
and exported together as a single FMI~3.0~\cite{FMI3:2023}
model-exchange FMU. The joint contributes the four continuous states
$(s,\dot s,w,\dot w)$, and the stacked heave and attitude joints a
further four position states and their derivatives. Two features of
this FMU matter for the optimal-control application. First, $\dot s$
is exported as the model variable \texttt{vsTrack}, which is the
factor \eqref{eq:sdot} through which the road geometry enters the
time-to-space transformation. FMIOPT reads it from the FMI interface
and rescales every other state derivative, with no rewriting of the
model. Second, the chassis coordinates are track-relative, so a
feasible initial guess for a full lap comes from setting every chassis
coordinate and every velocity except $\dot s$ to zero and prescribing
$s$ as a linear ramp from $0$ to $L$. That puts the vehicle on the
road surface with its axes aligned to the track triad at every
collocation point, without any preliminary simulation.

\section{Results}\label{sec:results}

The geometric quality of the surface fit was characterised in the
previous sections through the per-station residuals of the lateral
polynomial expansion (Section~\ref{sec:fit}) and the $C^4$
piecewise-quintic export (Section~\ref{sec:export}). We therefore
focus here on the downstream consequence of that geometry: whether the
resulting track model yields a well-conditioned minimum lap-time
problem. The representative case reported below is the Formula~1
Singapore circuit processed by the pipeline.

The \texttt{TrackJoint} component (Section~\ref{sec:trackjoint}) is
connected to the multi-body F1 vehicle model in Dymola, and the coupled
system is exported as a single FMI~3.0 FMU. FMIOPT then wraps this FMU
inside a CasADi~\cite{Andersson:2019} direct-collocation NLP and hands
the resulting problem to IPOPT~\cite{WaechterBiegler:2006} configured
with the MUMPS linear solver. The time-to-space
transformation of \eqref{eq:sdot} is supplied by \texttt{TrackJoint}
itself. The arc-length coordinate $s$ (\texttt{sTrack}) is a
continuous state of the joint and its time derivative $\dot s$ (\texttt{vsTrack}) is used in the formulation of the joint. FMIOPT uses
$\dot s$ to recast the dynamics from the time domain into the spatial
domain along the reference arc length $s$, so the conditioning of the
Jacobian factor $1-\kappa(s)\,w$ directly governs the convergence
behaviour of the resulting NLP.

The Singapore circuit contains corners whose radius is smaller than
the local road width, and the same geometry has appeared on several
other racing circuits processed by the pipeline. Because the centre-line parameterisation is
not even well defined on these circuits, the previous version of the
track parameterization had to introduce a geometric distortion of the reference by
adding a small lateral offset to the centre-line at the offending
stations, pushing the fold
$w_{\mathrm{fold}}(s)=1/\kappa_c(s)$ outside the track width. With the
minimum-curvature line no such distortion is required as, by construction,
$\min_s 1/|\kappa(s)|>b_{\max}$.

\begin{figure}[t]
\centering
\includegraphics[width=0.49\textwidth]{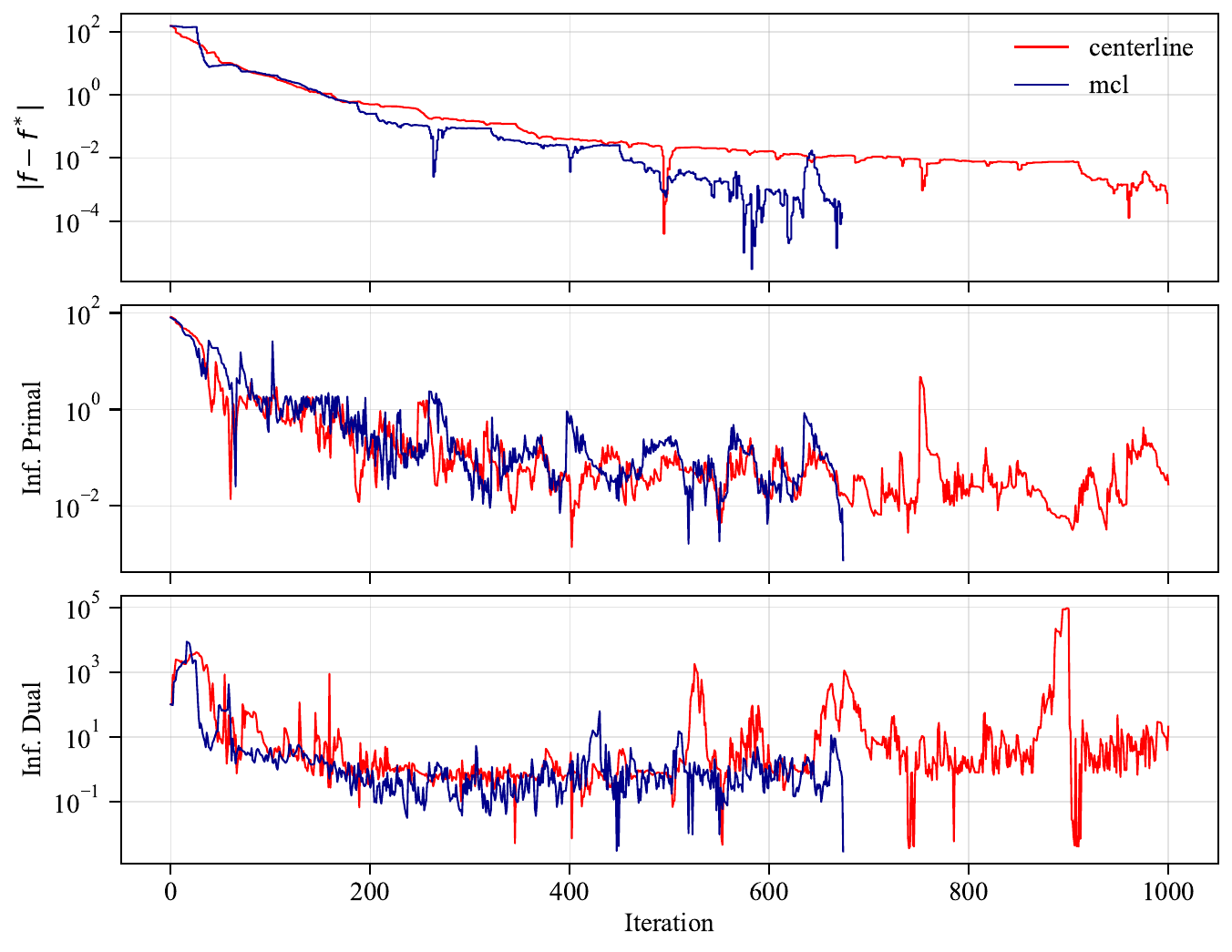}
\caption{IPOPT convergence trace of the minimum lap-time NLP on a
circuit containing a hairpin tighter than the road is wide. The
non-converging case uses the centre-line-anchored track joint of
\textcite{Limebeer:2023nascar}: the dual infeasibility jumps
erratically across iterations and the problem fails to converge.}
\label{fig:convergence}
\end{figure}

This geometric difference appears directly in the lap-time solve. As
shown in Fig.~\ref{fig:convergence}, the centre-line joint in the
ribbon-plus-Monge-patch form of \textcite{Limebeer:2023nascar} does
not converge: the dual infeasibility oscillates by several orders of
magnitude from one IPOPT iteration to the next, the primal residual
stalls and IPOPT adds regularisation in later iterations. This is
consistent with the numerical signature of an ill-conditioned KKT
system in the same tight-corner region where the centre-line
parameterisation is geometrically near its Frenet fold.

With the minimum-curvature joint, all other settings unchanged, the
Frenet fold is moved outside the drivable strip. Both primal and dual residuals decrease
monotonically to the IPOPT tolerances, and the NLP converges in a
comparable number of iterations to problems solved on
well-conditioned tracks. This makes several lap-time cases that were
previously intractable solvable without modifying the vehicle model or
solver settings.

\begin{figure}[t]
\centering
\includegraphics[width=0.49\textwidth]{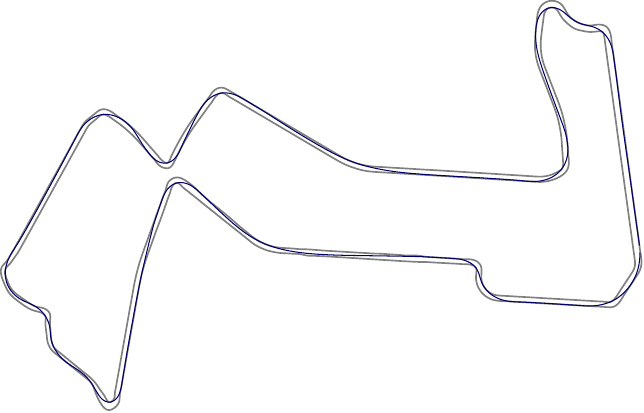}
\caption{Optimal racing line of the minimum lap-time solution on the
Singapore circuit, expressed as the lateral offset within the
drivable strip.}
\label{fig:racing-line}
\end{figure}

\begin{figure}[t]
\centering
\includegraphics[width=0.49\textwidth]{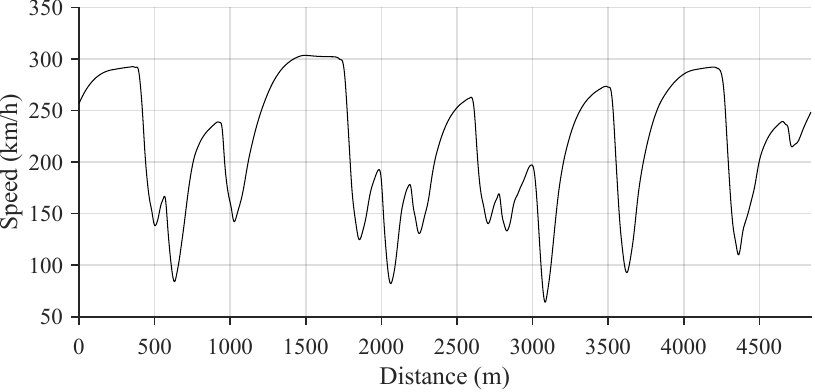}
\caption{Optimal speed profile of the minimum lap-time solution along
the reference arc length $s$.}
\label{fig:speed-profile}
\end{figure}

The optimised solution is shown in
Figs.~\ref{fig:racing-line} and~\ref{fig:speed-profile}. The racing
line of Fig.~\ref{fig:racing-line} traces the lateral offset $w(s)$
of the vehicle within the drivable strip, while the corresponding
speed profile of Fig.~\ref{fig:speed-profile} reports the optimal
velocity along the arc length $s$.

\section{Conclusions}

A complete pipeline has been described that turns a measured
racetrack point cloud into a closed, $C^4$-continuous Monge-patch
surface model for high-fidelity Modelica vehicle simulation. The main
point is that the centre-line reference used in
\cite{LimebeerVanDeMerwe:2025} and in earlier ribbon-based joints is
unsafe in any corner tighter than the road is wide. Anchoring the
Monge patch on the minimum-curvature line of
\textcite{Heilmeier:2020} and \textcite{Braghin:2008} removes the
Frenet singularity, improves the conditioning of the minimum lap-time
problem and gives a curvature profile close to the one the vehicle
actually experiences. The \texttt{TrackJoint} Modelica component
\cite{Sivaramakrishnan:2026joint} (Section~\ref{sec:trackjoint})
takes the exported piecewise-quintic splines and enforces
the two-degree-of-freedom kinematic constraint via d'Alembert's
principle, and it is the plug-in track interface for the FMIOPT
optimal-control toolchain. The export format consists of one knot
sequence shared by all
channels, quintic segments with $C^4$ continuity, and periodicity at
the wrap. The Modelica component consumes this data to construct the
track surface geometry and enforce the two-degree-of-freedom
kinematic constraint.
The pipeline runs in roughly 5~minutes per circuit and reproduces the
surveyed elevations and bounds to an order of a few centimetres.
Future work will look at replacing the rectangular $(s,n)$ grid with an
adaptive mesh, and at using a dynamic racing line as the reference to
make the time transformation more nearly linear.

\section*{Acknowledgements}

We gratefully acknowledge Joel Andersson for the development of the
FMIOPT toolchain that transcribes and solves the minimum lap-time
problems on which the pipeline is exercised.

\printbibliography

\end{document}